\documentclass[12pt]{article}

\usepackage{epsfig}

\begin{document}

\newlength{\oldparskip}
\setlength{\oldparskip}{\parskip} 
\oldparskip=\parskip 

\newcommand{\resetparskip}[0]
{
  \parskip=\oldparskip
}

\newcommand{\bit}[0]
{
  \parskip=-4pt
  \begin{itemize}
  \itemsep=-3pt
}
\newcommand{\eit}[0]
{
  \end{itemize}
  \parskip=-4pt
}

\newcommand{\ben}[0]
{
  \parskip=-4pt
  \begin{enumerate}
  \itemsep=-3pt
}
\newcommand{\een}[0]
{
  \end{enumerate}
  \parskip=-4pt
}

\newcommand{\bde}[0]
{
  \parskip=-4pt
  \begin{description}
  \itemsep=-3pt
}
\newcommand{\ede}[0]
{
  \end{description}
  \parskip=-4pt
}

\newcommand{\notes}[1]
{
  \mbox{ } \\
  NOTES
  \begin{verse}
  {#1}
  \end{verse}
  ENDNOTES \\
}

\newcommand{\tickle}[0]{Tcl/Tk }
\newcommand{\cpp}[0]{C[++] }

\newcommand{\needed}[2]{=={#1}=={#2}==}

\newcommand{\delete}[1]{}
\newcommand{\omitted}[1]{{\small \sf =OMITTED=}}

\renewcommand{\notes}[1]{}

\begin{titlepage}

\mbox{ } \\
\begin{figure}[!htbp]

  \centerline{\psfig{figure=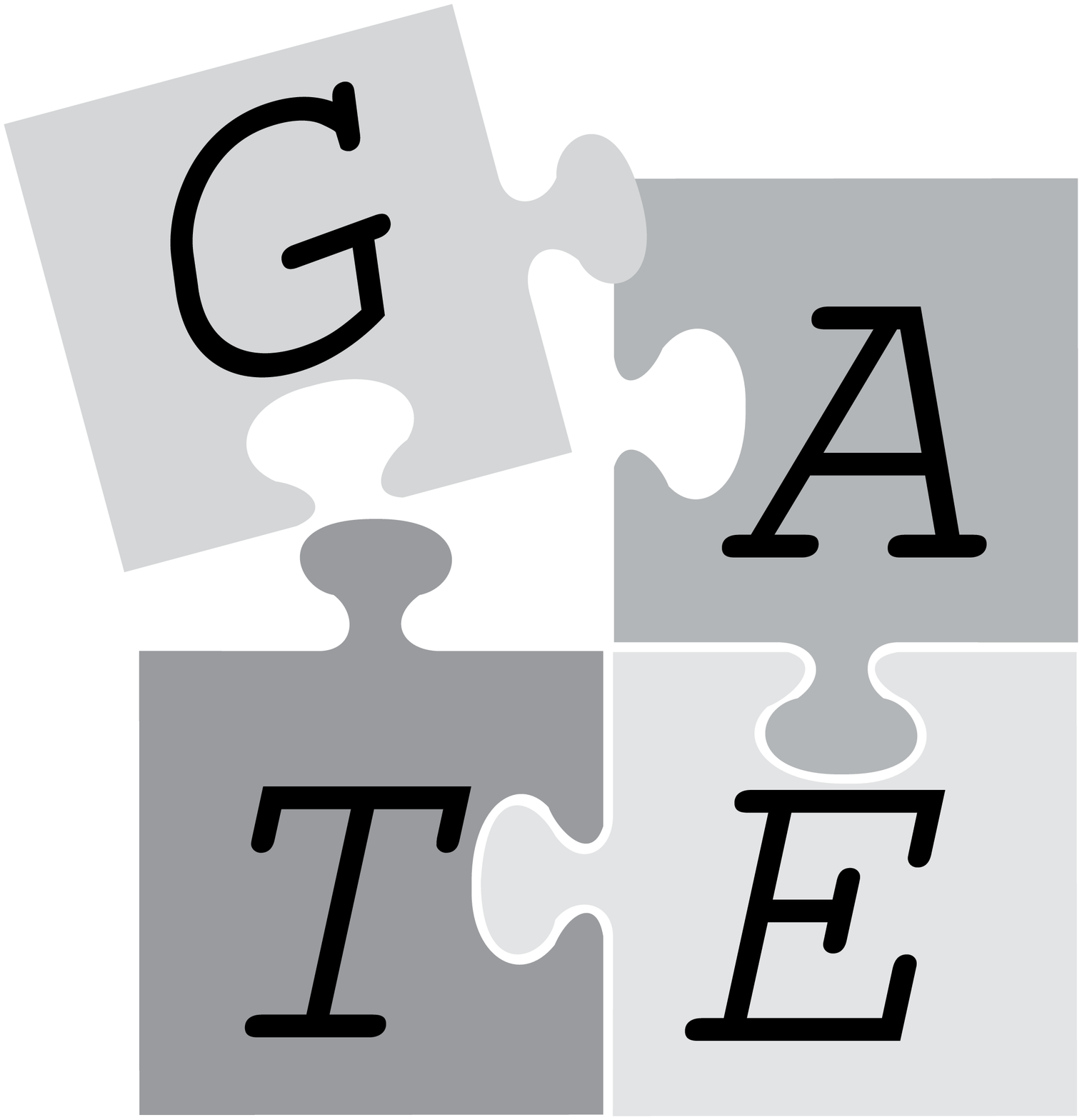,height=6cm}}
\end{figure}

\large
\vspace*{2in}
\begin{center}
\mbox{ } \\
\mbox{ } \\
\hspace*{1cm} CS -- 95 -- 21 \\
\mbox{ } \\
\hspace*{1cm} A General Architecture for Text \\
\hspace*{1cm} Engineering (GATE) -- a new approach to \\
\hspace*{1cm} Language Engineering R\&D \\
\mbox{ } \\
\hspace*{1cm} Hamish Cunningham \\
\hspace*{1cm} Robert J. Gaizauskas \\
\hspace*{1cm} Yorick Wilks

\end{center}
\end{titlepage}

\begin{titlepage}
\begin{center}

\Huge
A General Architecture for \\
Text Engineering (GATE) --- \\
a new approach to Language \\
Engineering R\&D
\begin{figure}[!h]

  \centerline{\psfig{figure=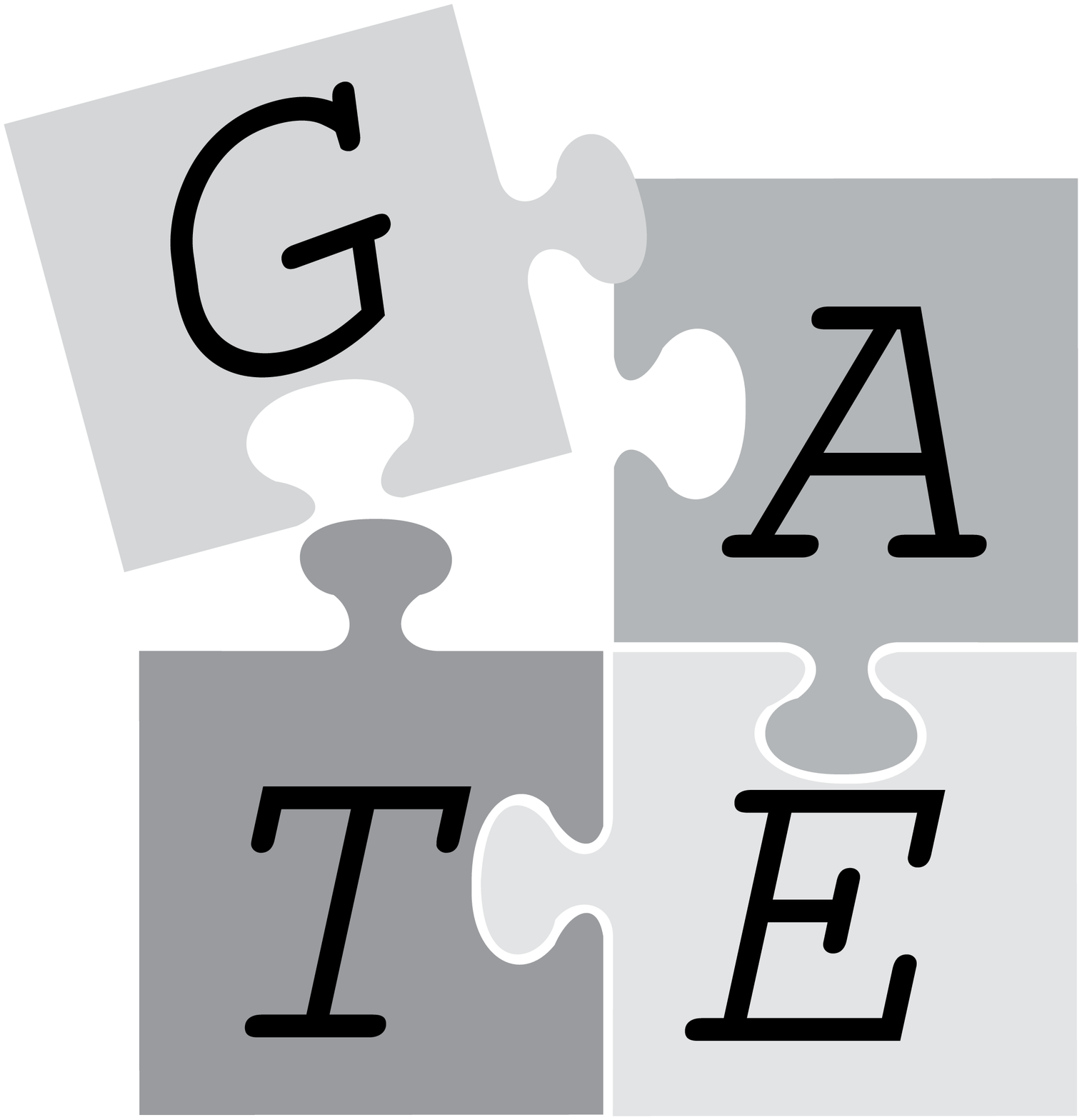,height=6cm}}
\end{figure}

\large
\vspace*{.5cm}
Hamish Cunningham \\
Robert J. Gaizauskas \\
Yorick Wilks \\

\mbox{ } \\
December 1995 \\
Research memo CS -- 95 -- 21 \\

\mbox{ } \\
Institute for Language, Speech and Hearing (ILASH), and \\
Department of Computer Science \\
University of Sheffield, UK \\

\mbox{ } \\
h.cunningham@dcs.shef.ac.uk \\
http://www.dcs.shef.ac.uk/research/groups/nlp/gate.html \\
http://www.dcs.shef.ac.uk/\verb|~|hamish

\end{center}
\end{titlepage}

\markboth{GATE -- tech. report CS - 95 - 21}
	 {GATE -- tech. report CS - 95 - 21}

\renewcommand{\baselinestretch}{0.5}
\small \normalsize
\tableofcontents
\renewcommand{\baselinestretch}{1.5}
\small \normalsize

\section{Introduction}

An increasing number of research and development
efforts have recently
positioned themselves under the banner {\em Language Engineering}
(LE). This signals a shift away from well-established labels such
as {\em Natural Language Processing} (NLP) and {\em Computational Linguistics}.
Examples include the renaming of UMIST's%
\footnote{The University of Manchester Institute of Science and Technology,
Manchester, UK}
{\em Department of Language and
Linguistics} (location of the {\em Centre for Computational Linguistics})
as the {\em Department of Language Engineering}, and the naming
of the European Commission's current relevant funding programme {\em
Language Engineering} (the previous programme was called {\em Linguistic
Research and Engineering}). The new journal of {\em Natural Language
Engineering} is another example%
\footnote{The editorial of the first issue also discusses the new name
(Boguraev, Garigliano, Tait 1995).}.

We shall argue here that this shift is more than simple TLA%
\footnote{TLA: three-letter acronym}%
-fatigue.
The new name reflects a change of emphasis within the field towards:

\bit
\item increasing use of quantitative evaluation as a metric of research
achievement;
\item renewed interest in statistical language models and
\linebreak
automatically-generated resources;
\item increasing availability and use of large-scale resources (e.g.
corpora, machine-readable dictionaries);
\item a re-orientation of language processing research to large-scale
applications, with a comcomitant emphasis on predictability and conformance
to requirements specifications (i.e. emphasis on engineering issues).
\eit

Section \ref{trends}
expands on these points, and the rest of the report then argues
that this shift requires
a more general approach to LE research and development, centred on the 
provision
of support software in the form of a general architecture and development
environment specifically designed for text processing systems. Under EPSRC%
\footnote{The Engineering and Physical Science Research Council, UK funding
body.}\
grant GR/K25267 the NLP group at the University of Sheffield are developing a
system that aims to implement this new approach (Wilks, Gaizauskas 1994).
The system is called GATE -- the General Architecture for Text Engineering.
\resetparskip

GATE is an {\em architecture} in the sense that it provides a common
infrastructure for building LE systems (rather like the frame of a building
or the interface specifications for the bus and peripherals of a computer). 
It is also a {\em development environment} that provides aids for the 
construction, testing and evaluation of LE systems (and particularly for the 
reuse of existing components in new systems).

Section \ref{gate} describes the architecture.
Section \ref{vie} describes an initial
application of GATE to collaborative research in Information Extraction (IE). 
Appendix \ref{related_work} discusses three existing systems:
\bit
\item
ALEP (Simpkins 1992), which turns out to be a rather different enterprise
from ours;
\item
MULTEXT (Thompson 1995a; Ballim 1995; Finch, Thompson, McKelvie 1995),
a different but largely
complementary approach to some of the problems addressed by GATE,
which is
particularly strong on SGML support and elements of which we intend to
integrate with GATE;
\item
TIPSTER (ARPA 1993a), whose architecture (Grishman 1995)
has been adopted as the storage
substructure of GATE, and which has been a primary influence in the design
and implementation of our system.
\eit
Appendix \ref{gate_design} is a
preliminary design and implementation document for GATE.
\resetparskip

\section{\label{trends}Current trends in Language Engineering R\&D}

We noted at the outset a recent trend towards
re-positioning language processing R\&D as {\em Language Engineering}.
This should not be taken to imply or require the death of [Computational]
Linguistics! The shift is quite possibly one of from theory to practice.
This section examines the background and consequences of the trend.

\subsection*{\label{toy_problem}Packing up the toys}

\notes{
  a re-orientation of language processing research on large-scale
  problems;

  ====
  quote magermann's toy problem syndrome, but note counterbalance of IBM group
  re-incorporating linguistic info (\& cite wilks article)
} 

Several commentators have characterised the
broad trend of AI approaches to language as tending towards the ``toy problem
syndrome'', expressing the view
that AI has too
often chosen to investigate artificial, small-scale
applications of the technology under development. These ``toy'' problems are
intended to be representative of the work involved in building applications of
the technology for end-user or ``real-world'' tasks, but
scaling up problem domains from the toy to the
useful has often shown the technology developed for the toy to be
unsuitable for the real job. 

For example,
one of the present authors began a large-scale Prolog grammar project in
1985 (Farwell, Wilks 1989):
by 1987 it was perhaps the largest DCG (Definite Clause Grammar)
grammar anywhere, designed to cover a linguistically well-motivated test set
of sentences in English. Interpreted by a standard parser it was able to
parse completely and uniquely
virtually no sentence chosen randomly from a newspaper. We
suspect most large grammars of that type and era did no better, though
reports are seldom written making this point.

The mystery for linguists is how that can be: the grammar appeared to
inspection to be virtually complete -- it {\em had} to cover English, if
thirty years of linguistic intuition and methodology had any value. It is a
measure of the total lack of evaluation of parsing projects up to that time
that such conflicts of result and intuition were possible, a situation
virtually unchanged since Kuno's large-scale Harvard parser of the
1960's (Kuno, Oettinger 1962) whose similar failure to produce a single,
preferred, spanning parse gave rise to the AI semantics and knowledge-based
movement. The situation was effectively unchanged in 1985 but the response
this time around has been quite different, characterised by:
\bit
\item
use of empirical methods with strict evaluation criteria;
\item
renewed interest in performance-based models of language, and a corresponding
renewal and extension of statistical techniques in the area;
\item
increased provision and reuse of large-scale data resources;
\item
greater emphasis on the
development of prototype applications of NLP technology to large-scale
problems.
\eit
\resetparskip

\subsection*{\label{quantitative}Measuring results with numbers} 

With hindsight it may seem obvious that {\em computational} linguistics, in
the sense of computer programs that seek to exploit the results of linguistic
research to make computers do useful things with human language%
\footnote{There is, of course, at least one 
other sense, that of using computational
tools to aid linguistic research.},
should be subject to empirical criteria of effectiveness. The big problem,
of course, is determining precisely what the criteria of success should be.
Should we collect video tapes of Star Trek and measure our efforts in
comparison to the Enterprise's lucid conversational computer? There is now a
substantial literature on this question (Crouch, Gaizauskas, Netter
1995; EAGLES 1994; Galliers, Sparck Jones 1993; Palmer, Finin 1990;
Sparck Jones 1994),
and more practical solutions to the
evaluation problem have emerged in a number of areas.

Participants in the TIPSTER programme and
the MUC (Message Understanding Conference, an information extraction
competition)
and TREC
(Text Retrieval Conference, an information retrieval (or `document
detection') competition)
competitions (ARPA 1992, 1994), for example,
build systems to do precisely-defined tasks on selected bodies of news
articles. Human analysts are employed to produce correct answers for some
set of previously unseen texts, and the systems run to produce machine
output for those texts. The performance of the systems relative to
human annotators is then measurable quantitatively.
Quantitative evaluation metrics bring numerically well-defined
concepts like precision and
recall, long used to evaluate information retrieval systems, to language
engineering%
\footnote{Machine Translation systems had, of course, always been subject to
rigourous evaluation from its earliest days (Lehrberger, Bourbeau 1988),
but this tradition did not spread further until recently.}%
.

It seems likely that the linking of quantitative performance metrics to
funding, as is the case in the U.S., has fostered a culture willing to
pursue any methods that are effective in these terms even where theoretical
purity suggests a different route. Whether this is a good or a bad thing
is left as an exercise for the reader. We note, however, that the recent
successes of speech recognition technology arose in a similar culture
(Church, Mercer 1993).

The U.S. model is not without significant disadvantages, however,
principally:
\ben
\item
a tendency to exclude novelty as sites all focus on one set of tasks;
\item
the high cost of producing evaluation data and administering competitive
evaluation.
\een
In the IE field (1) is evident in the current bias towards template-filling,
an application designed at the behest of the U.S. intelligence community.
\resetparskip

Regarding (2), analysis of the funding required for a European equivalent to
the American programmes has led the European Commission to reject
comparative evaluation (Cencioni 1995).

We shall argue in section \ref{gate} below that both of these problems can be
offset while retaining the benefits of empirical evaluation.

\subsection*{\label{quality}Performance vs. competence}

A related phenomenon is the increasing use of statistical techniques in the
field (Jelinek 1985; Church, Mercer 1993). Instead of an introspective
process of investigation into the underlying mechanisms by which people
process language (or, in Chomsky's terms, their {\em competence}),
statistical NLP attempts to build models of language as it exists in
practical use -- the {\em performance} of language. (Rens Bod's thesis
contains an extended discussion of this distinction -- Bod 1995.)

Statistical methods have had significant successes, and the debate once
thought closed by Chomsky's `I saw a fragile whale' is now as open as it
ever has been. Most part-of-speech taggers now rely on statistics
(Leech, Garside, Atwell 1983; Robert, Armstrong 1995)
and it seems possible that
parsers may also go this way (Church 1998; Magerman 1994, 1995;
Briscoe, Carroll 1993),
though more conventional methods are also increasing in quality and
robustness (Strzalkowski, Scheyen 1993).

It is possible that there is a natural ceiling to the advance of performance
models (Wilks 1994), but the point of relevance
for this report is that the jury is still out on performance vs. competence.
Thus, as well as a host of competing linguistic and lexicographic theories,
LE is home to a thoroughgoing
paradigm conflict. Two important consequences ensue.

First, empirical measurement of the relative
efficacy of competing techniques is even more important.
Secondly, hybrid models are becoming common, implying a growing
need for the flexible combination of different techniques in single
systems. Numbers of techniques that have poor performance alone may
sometimes be combined to produce a whole greater than the sum of the parts
(Wilks, Guthrie, Guthrie, Cowie 1992; Bartell, Cottrell, Belew 1994).

\subsection*{\label{reuse}Reusing resources in practice}

In common with other software systems, LE components deploy both data and
process elements. The quality, quantity and availability of shared data
resources has increased dramatically during the late 1980s and 1990s%
\footnote{Extensive discussion of the repositories (LDC, CLR, MLSR etc.) of
corpora and lexicon resources and their holidings up to 1994 can be found in
(Wilks et al. 1996). More recent developments concrening ELRA (the European
Linguistic Resources Association can be found in Elsnews 4.5 (November
1995).}%
.

The sharing of processing (or algorithmic)
resources remains more limited (Cunningham, Freeman, Black 1994),
one key reason being that the integration and reuse of different components
can be a major task. For example, the ESPRIT project PLUS devoted
substantial effort to reusing a theorem prover from IBM's STUF system for
parsing an HPSG grammar (Black ed. 1991). The COBALT project
(Rocca, Black, Cunningham, Zarri, Celnik 1993) failed to
locate a reusable shallow analysis engine with a cost-benefit profile for
reuse superior to reimplementation (Black, Cunningham 1993).
The CRISTAL project planned to reuse results from those projects
but again platform specificity had a negative impact (Cunningham, Underwood,
Black 1994).
Section \ref{trends} noted the increase in scale of the problems that
LE research
systems aim to tackle. In parallel with this trend, the overhead involved in
creating a full-scale IE system, for example, is also increasing. For many
research groups the costs are prohibitive. Any method for alleviating the
problems of reuse would make a signifcant contribution to LE research and
development.

On a smaller scale, the typical life-cycle of doctoral research in AI/NLP
is:
\bit
\item have an idea;
\item reinvent the wheel, fire and kitchen sinks to provide a framework for
the idea;
\item program the idea;
\item publish;
\item throw the system to the dogs / tape archivist / shelfware catalogue.
\eit
A framework which enabled relatively easy reuse of past work could
significantly increase research productivity in these cases.
\resetparskip

\subsection*{\label{engineering}Nuts and bolts}

With the increasing scale of LE systems, software engineering issues become
more important.
Just as the construction of the Severn Bridge was a rather different
order of problem from that of laying a couple of planks across a farmland
ditch, the development of software capable of processing megabytes of text,
written by idiosyncratic wetware%
\footnote{Journalists.},
in short periods of time to
measurable levels of accuracy is a quite different game from that, say,
of providing natural
language interaction for the control of a robot arm that moves
blocks on a table top (Winograd 1972).
The nuts and bolts are a lot bigger, and may even be of a
completely different fabric altogether.
This type of issue has been solved successfully in other areas of computer
science, e.g. databases. Failure to address software-level
robustness (as opposed to the robustness of the underlying NLP technology),
quality and efficiency will be a barrier to transferring LE technology from
the lab to marketplace.

Some other requirements relating to the technological foundations 
of these systems also arise.
Module interchangeability (at both the data and process levels), a kind of
`software lego' or `plug-and-play', would allow users to buy into LE
technology without tying them to one supplier. (In a different domain this
was the message of the Open Systems movement. Let's hope we don't emulate
their success!) Also desirable are easy upgrade routes as technology
improves.
In addition to the reasons noted above, precise quantification of performance 
measures are also needed to foster
confidence in the capability of LE applications to deliver, and robustness
and efficiency for large text volumes are prerequisites for many
applications. 
Software multilinguality and operating system independence are also issues.
Finally, maximising cross-domain portability will favourably
impact delivery costs. (See (Grishman 1995) for a similar review of these
points.)

\subsection*{\label{super_highway}Gridlock on the super-highway}

\notes{
  ever-growing needs for language processing applications that do more
  than treat texts as unordered bags of words.

  ==== trad text processing = IR = unordered sets of stemmed words (not to say
  that IE is dead, far from it...); internet growth stats from MIIS book etc.;
  URLs for companies on the web)

  see orig gate docs
} 

Our discussion of trends in LE concludes with 
two major LE application areas, Information Extraction (IE) and Machine
Translation (MT), which both exhibit the trends discussed above.

Recent years have seen significant improvements in the quality and
robustness of LE technology.
Rapid improvement in robustness (the ability to deal with any input) and
quality are evident in the leading systems
(Jacobs ed. 1992; ARPA 1992; ARPA 1993b; Strzalkowski, Scheyen 1993;
Magerman 1995).
In this year's MUC-6 competition
(Sundheim 1995a,b; Onyshkevych 1995a,b)
initial results indicate that named-entity recognition can now
be performed by machines to performance levels equal those of
people (ARPA 1996; Wakao, Gaizauskas, 1995).
The result is that
applications of the technology to large-scale problems are increasingly
viable.

IE is intended to deal with the
rapidly growing problem of extracting meaningful information from
the vast amount of electronic textual data that threatens to engulf us.
Scientific journal abstracts, financial newswires, patents and patent
abstracts, corporate and government technical documentation, electronic mail
and electronic bulletin boards all contain a wealth of information of vital
economic, social, scientific and technical importance. The problem is that the
sheer volume of these sources is increasingly preventing the timely finding
of relevant information, a state of affairs exacerbated by the
explosive growth of the Internet (Kroll 1994; Thompson 1995b).

Existing information retrieval (Salton 1989) solutions to this problem  are
a step in the right direction, and the industry supplying IR applications
can expect to continue in its current healthy state.

IR systems, however, attempt no analysis of the meaningful content of
texts.
This is a
strength of the approach, leading to robustness and speed, but also a
weakness, as the information represented by the texts is retrieved in the
format of the texts themselves -- i.e. in the ambiguous and verbose medium
of natural language.

Extraction of information in definite formats is an obvious solution and one
which can only be achieved through the application of LE technology.

The IE community have been leaders in quantitative evaluation (ARPA 1993b).
Statistical methods are widely used, but so is more conventional CL
(ARPA 1993).
The need for systematic reuse
of both data and processing resources has been recognised, and work funded
to facilitate this, and the importance of software engineering matters noted
(TIPSTER 1994).

A similar situation is evident in MT research. Nyberg, Frederking, Farwell,
Wilks (1994) note the
continuing importance of evaluation for MT; Nirenburg, Mitamura, Carbonell
(1994) propose
that the multi-approach, multi-paradigm nature of the field be embodied in
`multi-engine', or `adaptive' MT systems.

\section{\label{gate}GATE}

\notes{

summary of motivation and requirements from above

philosophy wrt MULTEXT, TIPSTER, ALEP: yes MULTEXT (with reservations); yes
TIPSTER (with additions); no ALEP (in the sense that GATE is NOT `a system
for doing X' but a backplane for building systems to do X, Y, Z...)

the 3 elements

high-level architecture diagram(s); screendumps

development plan (decompose \& reconstitute LaSIE, OO tech (=shopping list);
...)

Any particular delivery or task-specific instantiation of GATE will include
some subset of CREOLE. The first distribution will include VIE, a set of
CREOLE objects based on LaSIE.

} 

The previous section argued that the new name, {\em Language Engineering},
reflects changes of territory for natural language R\&D, and drew out a set
of requirements for LE systems. We believe that these requirements may best
be met by the provision of dedicated support software for researchers and
applications developers, in the form of an {\em architecture} and {\em
development environment}, and we have developed an initial version of such a
system, called GATE - a General Architecture for Text Engineering.

GATE is an {\em architecture} in the sense of providing a common
infrastructure for building LE systems. An analogy is the hardware
architecture of a PC: provided a manufacturer of, say multi-media
controller cards follows the published specification of the PC bus, BIOS
etc., the card should work in any machine. Further, the card should be able
to rely on certain common services provided by the PC architecture.

GATE is a {\em development environment} because it provides a variety of
data visualisation, debugging and evaluation tools (with point-and-click
interface), and a set of standardised interfaces to reusable components. It
supports the development of LE systems in a way analogous to the support for
program development provided by compilers, libraries, debuggers and
syntax-aware editors.

GATE will be available free for research purposes, and is intended to grow
and develop in response to the needs of the UK and European LE communities.
Its design incorporates results from related European and US initiatives
and bridges the infrastructural work of the two.

The rest of this section:
\bit
\item
casts the constraints on LE systems identified in section \ref{trends} as
requirements for GATE;
\item
gives an overview of the architecture in the context of the requirements;
\item
describes the arrangements for collaborative work on IE using the initial
distribution of GATE;
\item
gives a roadmap of future development of the system.
\eit
More detail on GATE can be found in Appendix \ref{gate_design}, and on
related work in Appendix \ref{related_work}.
\resetparskip

\subsection*{Summary of requirements}

A general architecture for LE R\&D should:
\bit
\item support collaborative research;
\item support hybrid systems, `plug-and-play' module interchangeability,
and easy upgrading;
\item
support the reuse of existing and future algorithmic components and data 
resources, whether they be the results of PhD projects or multinational
strategic initiatives;
\item
contribute to software-level robustness, quality and efficiency;
\item
contribute to portability across problem domains and application areas;
\item
support comparative evaluation, preferably at lower cost than the
US programmes and without stifling innovation;
\item
contribute to software portability across languages and across operating
systems and programming languages.
\eit
\resetparskip

\subsection*{Architecture overview}

GATE presents LE researchers or developers with an environment where they
can use tools and linguistic databases easily and in combination, launch
different processes, say taggers or parsers, on the same text and compare
the results, or, conversely, run the same module on different text
collections and analyse the differences, all in a user-friendly interface.
Alternatively, module sets can be strung together to make e.g. IE, IR or MT
systems. Modules and systems can be evaluated (using e.g. the Parseval tools),
reconfigured and reevaluated -- a kind of edit/compile/test cycle for LE
components.

\begin{figure}[!htbp]
  \centerline{\psfig{figure=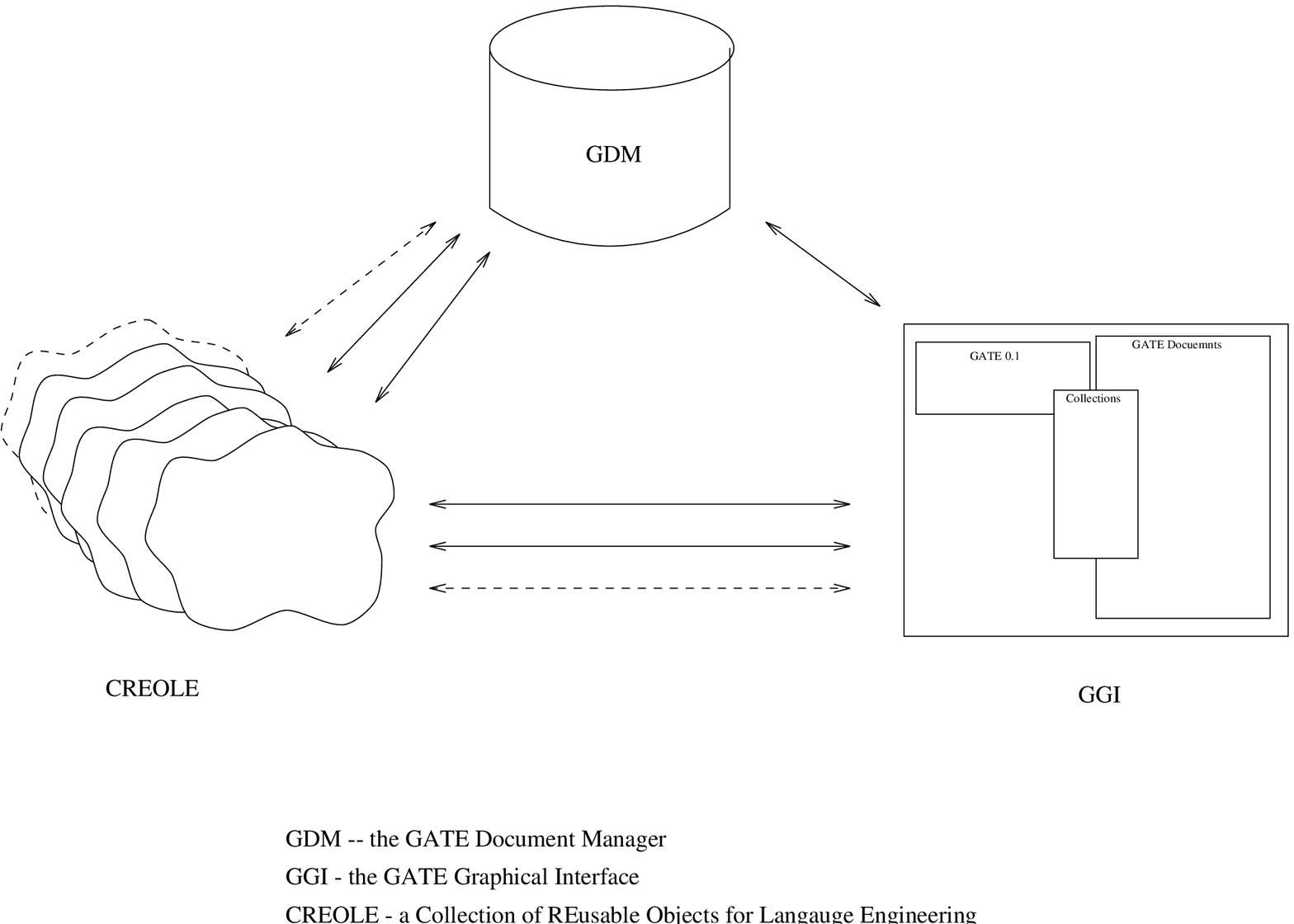,height=8cm}}
  \caption{\label{arch1}The three elements of GATE}
\end{figure}
GATE comprises three principal elements (figure \ref{arch1}):
\bit
\item
a database for storing information about texts and a database schema
based on an object-oriented model of information about texts (the GATE
Document Manager -- GDM);
\item
a graphical interface for launching processing tools on data and viewing and
evaluating the results (the GATE Graphical Interface -- GGI);
\item
a collection of wrappers for algorithmic and data resources that
interoperate with the database and interface and constitute a Collection of
REusable Objects for Language Engineering -- CREOLE.
\eit
\resetparskip

GDM is based on the TIPSTER document manager, and the initial implementation
supplied by the Computing Research Lab at New Mexico State University (whose
help we gratefully acknowledge). It is planned to enhance the SGML
capabilities of this model, possibly
by exploiting the results of the MULTEXT project
(we thank colleagues from ISSCO and Edinburgh for making available
documentation and advice on this subject).
See Appendix \ref{related_work} for details of the relationship between GATE
and these (and other) projects.

GDM provides a central repository or server that stores all information an
LE system generates about the texts it processes. All communication between
the components of an LE system goes through GDM, insulating parts from each
other and providing a uniform API (applications programmer interface)
for manipulating the data produced by the
system.%
\footnote{Where very large data sets need passing between modules other
external databases can be employed if necessary.}\
Benefits of this approach include the ability to exploit the maturity and
efficiency of database technology, easy modelling of blackboard-type
distributed control regimes (of the type proposed by: Boitet, Seligman 1994;
section on control in Black ed. 1991) 
and reduced interdependence of components.

GGI is in development at Sheffield. It is a graphical launchpad for LE
subsystems, and provides various facilities for viewing and testing results
and playing software lego with LE components: interactively stringing
objects into different system configurations.

All the real work of analysing texts (and maybe producing summaries of them,
or translations, or SQL statements\ldots) in a GATE-based LE system is done
by CREOLE modules.

Note that we use the terms {\em module} and {\em object} rather loosely to mean
interfaces to resources which may be predominantly algorithmic or
predominantly data, or a mixture of both. We exploit object-orientation for
reasons of modularity, coupling and cohesion, fluency of modelling and ease
of reuse (see e.g. Booch 1994).

Typically, a CREOLE object will be a wrapper around a pre-existing LE module
or database -- a tagger or parser, a lexicon or ngram index, for example.
Alternatively objects may be developed from scratch for the architecture --
in either case the object provides a standardised API 
to the underlying resources which allows access via GGI and
I/O via GDM. The CREOLE APIs may also be used for programming new objects.

The initial release of GATE will be delivered with a CREOLE set comprising a
complete MUC-compatible IE system (to begin with,
more of a pidgin than a creole!). Some of the objects will be based on
freely available software (e.g. the Brill tagger (Brill 1994)),
while  others are derived from Sheffield's MUC-6 entrant, LaSIE%
\footnote{Large-Scale IE.}
(Gaizauskas, Humphreys, Wakao, Cunningham 1995;
Gaizauskas, Humphreys, Wakao, Cunningham 1996).
This set is called VIE -- a Vanilla IE system. See
section \ref{vie} for an overview.
CREOLE will expand quite rapidly during 1996, to
cover a wide range of LE R\&D components (such as those currently available
at the ACL-sponsored Natural Language Software Registry at
DFKI%
\footnote{URL: http://cl-www.dfki.uni-sb.de/cl/registry/draft.html}%
), but for the rest of this
section we'll use IE as an example of the intended operation of GATE.

The recent MUC competition, the sixth, defined four IE tasks to be carried out
on Wall Street Journal articles. Sheffield's system did well, scoring in the
middle of the pack in general
and doing as well as the best systems in some areas.
Developing this system took 24 person-months, one significant element of which
was coping with the strict MUC output specifications. What does a research
group do which either does not have the resources to build such a large system,
or even if it did would not want to spend effort on areas of language
processing outside of its particular specialism? The answer until now has
been that these groups cannot take part in large-scale system building, thus
missing out on the chance to test their technology in an
application-oriented environment
and, perhaps more seriously, missing out on the extensive quantitative
evaluation mechanisms developed in areas such as MUC. In GATE and VIE we
hope to provide an environment where groups can mix and match elements of
MUC technology from other sites (including ours)
with components of their own, thus allowing the benefits
of large-scale systems without the overheads. A parser developer, for
example, can replace the parser supplied with VIE.

Licencing restrictions
preclude the distribution of MUC scoring tools with GATE, but Sheffield will
arrange for evaluation of data produced by other sites.
In this way, GATE/VIE will support comparative evaluation of LE components
at a lower cost than the ARPA programme (partly by exploiting their
work, of course!). Because of the relative informality of these evaluation
arrangements, and as the range of evaluation facilities in GATE expands
beyond the four IE tasks of the current MUC, we should also be able to offset
the tendency of evaluation programmes to dampen innovation.

Similarly we aim to make collaboration between research groups much easier.
Sites specialising on different LE
subtasks can combine their efforts into bigger application-oriented systems
with minimal overhead. We hope that we can help the community squeeze a
little more research time out of industrially-oriented projects by cutting
down on the time spent integrating research work into demonstrator systems.

Working with GATE/VIE, the researcher will from the outset reuse existing
components, the overhead for doing so being much lower than is
conventionally the case -- instead of learning new tricks for each module
reused, the common APIs of GDM and CREOLE mean only one integration
mechanism must be learnt. And as CREOLE expands, more and more modules and
databases will be available at low cost. We also endorse object orientation
(OO) in this context, as an enabling technology for reuse (Booch 1994),
and hope to move towards sub-component
level reuse at some future point, possibly
providing C++ libraries as part of an OO LE
framework (Cunningham, Freeman, Black 1994).

As we built our MUC system it was often the case that we were unsure of the
implications for system performance of using tagger X instead of tagger Y,
or gazeteer A instead of pattern matcher B. In GATE, substitution of
components is a point-and-click operation in the GGI interface. (Note that
delivered systems, e.g. EC project demonstrators, can use GDM and CREOLE
without GGI -- see below.) This facility supports hybrid systems, ease of
upgrading and open systems-style module interchangeability.

Of course, GATE does not solve all the problems involved in plugging diverse
LE modules together. There are two barriers to such integration:
\bit
\item
incompatability of {\em representation} of information about text and the
mechanisms for storage, retrieval and inter-module communication of that
information;
\item
incompatability of {\em type} of information used and produced by different
modules.
\eit
GATE enforces a separation between these two and provides a solution to the
former based on the work of the TIPSTER architecture group (TIPSTER 1994).
Because GATE places no constraints on the linguistic formalisms or
information content used by CREOLE objects, the latter problem must be solved 
by dedicated translation functions -- e.g. tagset-to-tagset mapping -- and, in
some cases, by extra processing -- e.g. adding a semantic processor to
complement a bracketing parser in order to produce logical form to drive a
discourse interpreter. As more of this work is done we can expect the
overhead involved to fall, as all results will be available as CREOLE
objects. In the early stages Sheffield will provide some 
resources for this work in order to get the ball rolling, i.e. we will
provide help with CREOLEising existing systems and with developing interface
routines where practical and necessary. We are confident that integration
{\em is} possible (partly because we believe that differences between
representation formalisms tend to be exaggerated) -- and others share this
view, e.g. the MICROKOSMOS project (Beale, Nirenburg, Mahesh 1995), which
seeks to integrate many types of knowledge source in a useable whole, as
well as the LexiCadCam experience at New Mexico (Wilks, Guthrie,
Slator 1996) which
sought to provide core lexical information as needed in a range of
user-specified formats.
\resetparskip

\begin{figure}[!htbp]
  \centerline{\psfig{figure=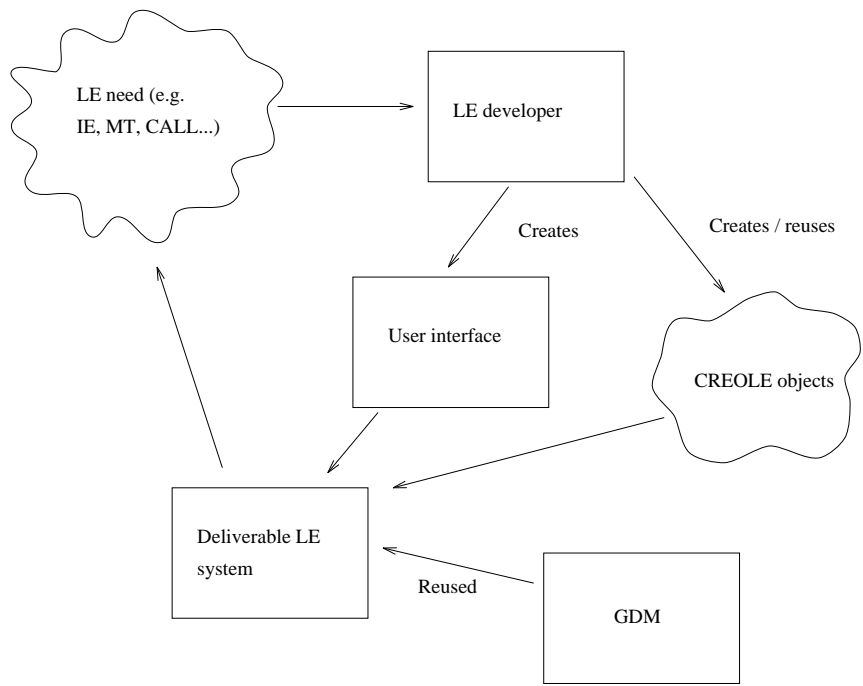,height=9cm}}
  \caption{\label{delivery}Delivering systems with GATE}
\end{figure}

GATE is also intended to benefit the LE system developer (which may be the
LE researcher with a different hat on, or industrialists implementing
systems for sale or for their own text processing needs).
Using GATE for the delivery of a system is illustrated in figure
\ref{delivery}.
A delivered system comprises a set of CREOLE objects, the GATE runtime
engine (GDM and associated APIs) and a custom-built interface (maybe just
character streams, maybe a Visual Basic Windows GUI, \ldots). The interface
might reuse code from GGI, or might be developed from scratch.

The LE user may upgrade by swapping parts of the CREOLE
set if better technology becomes available elsewhere. This model for the
commercialisation of LE technology is already begininning to operate in the
US, where a number of organisations are preparing TIPSTER-compatible modules
for sale or distribution for research.
(These organisations include NMSU, SRA, HNC,
University of Massachusetts, Paracell, Logicon (Dunning 1995, personal
communication).) All TIPSTER-compatibile modules will also work with GATE as
GATE itself is desinged to be a TIPSTER-compatible system. Thus the pool of
easily reusable LE resources available to researchers and developers using
GATE has the potential to become a large, rich set of modules from a good
proportion of the LE community world-wide. Also, it may well become the case
that organisations purchasing LE software will require TIPSTER compatability
(this will be true of US government organisations, for example).

At the software engineering level GATE:
\bit
\item
contributes to robustness and quality by providing a mature infrastructure;
\item
contributes to efficiency via the design of the TIPSTER text model and by
access to fast database technology underlying GDM.
\eit
\resetparskip

As regards operating system independence, GDM and GGI will initially be
available for Linux, SunOS, Solaris 2 and other UNIX platforms as required,
but will avoid using UNIX-specific facilities. A Windows version may
follow at some point (see the roadmap section below). CREOLE portability is
more difficult. GATE places no constraints on the implementation languages
and platforms of CREOLE objects, so they may or may not be portable.

GATE cannot eliminate the overheads involved with porting LE systems to
different domains (e.g. from financial news to medical reports). Tuning LE
system resources to new domains is a current research issue
(see also: the LRE DELIS and ECRAN projects; Evans, Kilgariff 1995).
The modularity of
GATE-based systems should, however, contribute to cutting the engineering
overhead involved.

\subsection*{Collaboration using GATE and VIE}

Sheffield will support collaborative work using GATE/VIE for LE research
groups (typically academic groups), businesses with IE needs and
producers of lexicons and dictionaries.  The three groups are
{\em technology}, {\em data} and {\em resource providers} respectively,
contributing CREOLE modules, test data (e.g. manually extracted information
and the relevant source texts) and machine-readable language resources (e.g.
dictionaries).
The projected benefits for participants include:
\bit
\item
comparative quantitative evaluation of candidate technologies for IE;
\item
technology providers
may specialise on components of the IE task, avoiding the overhead of
providing a complete IE system while still working within the framework of
a complete NLP application;
\item
data providers (typically industrial concerns) get access to  IE technology
applied to their particular textual problem domains;
\item
resource providers can assess the performance of their products and increase
the market for them by encouraging their use in applied LE systems.
\eit
Note that there will be no requirement to supply source code for contributed
modules, and that intellectual property and other rights will be safeguarded
by appropriate legal agreements.
\resetparskip

\subsection*{Roadmap}

\notes{
  IE shell (with VIE)

  MT etc. (IR; CALL; multimedia, vision, ... speech?)

  resource repository
} 

Our first goal for GATE is to provide a prototype
of the architecture along with a set of CREOLE objects for doing MUC-style
information extraction. 
GATE/VIE 1.0 will be available at the start of 1996
to research groups and development projects who wish to participate in IE
systems development. We will at that point solicit contributions of CREOLE
replacements for VIE modules, and data sets from organisations with IE
needs.

Initial versions will run under UNIX and X11 only, and support the gcc
\cpp compilers and Tcl 7.4 / Tk 4.0 (Ousterhoot 1994) and higher.

Subsequent developments will concentrate on expanding the set of CREOLE
objects in order to:
\bit
\item exemplify the use of GATE in other application areas,
e.g. MT, CALL, Speech research;
\item make GATE a standard resource repository via CREOLE wrappers for LE
resources like lexicons, grammars etc. (possibly in collaboration with the
newly-formed European Language Resources Association).
\eit
Sheffield will contribute resources to integration of other sites'
components to start with.
\resetparskip

On the technical side issues include:
\bit
\item 16 bit character support;
\item internationalisation of system messages to increase ease of use
for non-native English speakers (probably taking into account the results of
LRE project 61-003 GLOSSASOFT (Hudson 1995));
\item evaluation and revision of the GGI interface, and the addition of
further data visualisation and debugging facilities;
\item SGML support;
\item portability to other platforms.
\eit
\resetparskip

We envisage considerable input from other research groups and welcome
crticism and comment on our implementation (and offers of work!).

\section{\label{vie}VIE, a Vanilla Information Extraction system}

\notes{

application of NE recognition to IR precision:

you wouldn't want to specify all the patterns for e.g. Tues 3rd May 1965 -
instead want to specify logical information

} 

As originally envisaged (Wilks, Gaizauskas 1994), GATE will be distributed
with a set of CREOLE objects that together implement a complete information
extraction system capable of producing results compatible with the MUC-6
task definitions. This CREOLE set is called VIE, a Vanilla IE
system, and it is intended that participating sites use VIE as the basis for
specialising on sub-tasks in IE. By replacing a particular VIE module -- the
parser, for example -- a participating group will immediately be able
to evaluate their specialist technology's potential contribution to
full-scale IE applications. Sheffield has access to the MUC-6 scoring tools
(and the PARSEVAL software) and will run periodic evaluations of various
VIE-based configurations.

It is envisaged that LE research groups (typically academic research
groups) supply modules
to replace parts of VIE. Businesses with IE needs can also participate in
the programme by contributing test sets and task definitions.
Resource builders like dictionary
publishers will be approached to supply research versions of their online
texts.

The most recent MUC competition, MUC-6, defined four tasks to be carried
out on Wall Street Journal articles:
\bit
\item
named entity (NE) recognition, the recognition and classification of definite
entities such as names, dates, places;
\item
coreference (CO) resolution, the identification of identity relations
between entities (including anaphoric references to them);
\item
template element (TE) construction, a fixed-format, database-like
enumeration of organisations and persons;
\item
scenario template (ST) construction, the detection of specific relations
holding between template elements
relevant to a particular information need (in this case personnel joining
and leaving companies) and construction of a fixed-format structure
recording the entities and details of the relation.
\eit
VIE is an integrated system that builds up a single, rich model of a text
which is then used to produce outputs for all four of the MUC-6 tasks. Of
course this model may also be used for other purposes aside from MUC-6
results
generation, for example we currently generate natural language summaries of
the MUC-6 scenario results.
\resetparskip

Put most broadly, and superficially, VIE's approach involves compositionally
constructing semantic representations of individual sentences in a text
according to semantic rules attached to phrase structure constituents which
have been obtained by syntactic parsing using a corpus-derived context-free
grammar. The semantic representations of successive sentences are then
integrated into a `discourse model' which, once the entire text has been
processed, may be viewed as a specialisation of a general world model with
which the system sets out to process each text.

Features which distinguish the system are:

\bit
\item an integrated approach allowing knowledge at several linguistic
  levels to be applied to each MUC-6 task (e.g. coreference
  information is used in named entity recognition);
\item the absence of any overt lexicon -- lexical information needed
  for parsing is computed dynamically through part-of-speech-tagging
  and morphological analysis;
\item the use of a grammar derived semi-automatically from the Penn
  TreeBank corpus;
\item the use and dynamic acquisition of a world model, in particular for
the
  coreference and scenario tasks;
\item a summarisation module which produces a brief natural language
  summary of scenario events.
\eit
VIE will be available with the initial release of GATE.
\resetparskip

\section{\label{conclusion}Conclusion}

We have argued that the language processing field is in a state of rapid
change, and that the
focus is shifting to large-scale applications and systems that are beginning 
to produce marketable solutions. The new emphasis has generated a new name --
{\em Language Engineering}.

We suggest that a new approach to software support for LE R\&D should be
developed to parallel this shift. We have proposed an architectural solution
-- GATE -- grounded on previous work in the area.

GATE aims to be a standard architecture for LE systems. Standards, of
course, must sell themselves -- imposition rarely works (whether in
computer science or in real life!).%
\footnote{One of a legion of examples is the relative popularity of TCP/IP
and OSI in the networking world.}
We hope that the LE community will endorse our argument for an LE support
architecture, and that our implementation will be strong enough to fulfil the
promise of the idea.

\section{\label{references}References}

(ARPA 1992) Advanced Research Projects Agency,
{\em Proceedings of the Fourth Message Understanding Conference (MUC-4)},
Morgan Kaufmann, 1992.

(ARPA 1993a) Advanced Research Projects Agency,
{\em Proceedings of TIPSTER Text Program (Phase I)}
Morgan Kaufmann, 1993.

(ARPA 1993b) Advanced Research Projects Agency,
{\em Proceedings of the Fifth Message Understanding Conference (MUC-5)},
Morgan Kaufmann, 1993.

(ARPA 1996) Advanced Research Projects Agency,
{\em Proceedings of the Sixth Message Understanding Conference (MUC-6)},
Morgan Kaufmann, 1996.

(Ballim 1995) A. Ballim, {\em Abstract Data Types for MULTEXT Tool I/O},
LRE 62-050 Deliverable 1.2.1, 1995.

(Bartell, Cottrell, Belew 1994) B.T. Bartell, G.W. Cottrell, R.K. Belew,
{\em Automatic Combination of Multiple Ranked Retrieval Systems},
proceedings of SIGIR 1994.

(Beale, Nirenburg, Mahesh 1995)
S. Beale, S. Nirenburg, K. Mahesh.
{\em Semantic Analysis in
the Mikrokosmos Machine Translation Project},
{\em Proceedings of the Second Symposium on Natural Language Processing
(SNLP-95)}, August 2-4, 1995, Bangkok, Thailand.

(Black ed. 1991) W.J. Black (ed.), {\em PLUS -- a Pragmatics-Based Language
Understanding System, Functional Design}, ESPRIT P5254 Deliverable D1.2,
1991.

(Black, Cunningham 1993)  W.J. Black, H. Cunningham,
{\em SCAM - The Surface-level COBALT Analysis Module},
LRE 61-011 deliverable, and Technical Report C/CCL/04/93/a, Centre for
Computational Linguistics, UMIST, 1993.

(Bod 1995) R. Bod, {\em Enriching Linguistics with Statistics: Performance
Models of Natural Language}, PhD thesis,
Institute for Logic, Language and Computation,
University of Amsterdam 1995.

(Boguraev, Garigliano, Tait 1995), B. Boguraev, R. Garigliano, J. Tait,
editorial of {\em Journal of Natural Language Engineering, Vol. 1 Part 1},
Cambridge University Press, March 1995.

(Boitet, Seligman 1994) C. Boitet, M. Seligman, {\em The ``Whiteboard''
Architecture: A Way to Integrate Heterogeneous Components of NLP Systems},
proceedings of COLING 1994.

(Booch 1994) G. Booch, Object-oriented Analysis and Design 2nd. Edtn.,
Addison Wesley 1994.

(Brill 1994) E. Brill,
{\em Some Advances in Transformation-Based Part of Speech Tagging},
proceedings of The Twelfth National Conference on Artificial Intelligence
(AAAI-94), Seattle, Washington 1994.

(Briscoe, Carroll 1993) T. Briscoe, J. Carroll, {\em Generalized
Probabilistic LR Parsing of Natural Language (Corpora) with
Unification-Based Grammars}, 
{\em Computational Linguistics} Volume 19, Number 1, March 1993.

(Cencioni 1995) R. Cencioni, speaking at the
plenary session of {\em The Second Language
Engineering Convention}, London 1995.

(Church 1988) K. Church, {\em A stochastic parts of speech program
and noun phrase parser for unrestricted text}, {\em Second
conference on Applied NLP}, ACL 1988.

(Church, Mercer 1993) K.W. Church, R.L. Mercer, {\em Introduction to the
Special Issue on Computational Linguistics Using Large Corpora},
{\em Computational Linguistics} Volume 19, Number 1, March 1993.

(Crouch, Gaizauskas, Netter 1995) R. Crouch, R. Gaizauskas, and K. Netter, 
{\em Interim Report of the Study Group on Assessment and Evaluation}, Report
prepared under the auspices of the EAGLES Project for the
Language Engineering Sector, Telematics Programme.
To be published shortly.

(Cunningham 1994)
H. Cunningham, {\em Support Software for Language Engineering
Research}, Technical Report 94/05, Centre
for Computational Linguistics, UMIST, Manchester 1994.

(Cunningham, Underwood, Black 1994)
H. Cunningham, N.L. Underwood, W.J. Black, {\em Reusable Components
for CRISTAL}, LRE Project 62-059 Technical Report, Centre  
for Computational Linguistics, UMIST, Manchester 1994.

(Cunningham, Freeman, Black 1994)
H. Cunningham, M. Freeman, W.J. Black, {\em Software Reuse, 
Object-Orientated Frameworks and Natural Language Processing}, 
Conference on New Methods in Natural Language Processing, 
Manchester 1994.

(EAGLES 1994) {\em Evaluation of Natural Language Processing Systems}, draft
document EAG-EWG-IR.2 of the EAGLES Project for the
Language Engineering Sector, Telematics Programme,
Fourth Framework Programme of the European Commission.

(Evans, Kilgarriff 1995) R. Evans, A. Kilgarriff, {\em Standards 
and How to do Lexical Engineering},
{\em Proceedings of the Second Language Engineering
Convention}, London 1995.

(Farwell, Wilks 1989) D. Farwell, Y. Wilks, {\em ULTRA -- a
multilingual machine translation system}, memoranda in Computer and
Cognitive Science, CRL Las Cruces, NM 1989.

(Finch, Thompson, McKelvie 1995) S. Finch, H. Thompson, D. McKelvie,
{\em Specification of Tool Shell with Discussion of Data and Process
Architecture},
LRE 62-050 Deliverable 1.2.2, 1995.

(Fr\"{o}lich, Werner 1994) M. Fr\"{o}lich, M. Werner, {\em The Graph
Visualisation System daVinci -- A User Interface for Applications},
technical report 5/94, Dept. Computer Science, University of Bremen 1994.

(Gaizauskas, Cahill, Evans 1993) 
R. Gaizauskas, L.J. Cahill, R. Evans,
{\em Description of the Sussex System Used for (MUC-5)},
in {\em Proceedings of the Fifth Message Undersanding Conference (MUC-5)},
ARPA, Morgan Kaufmann, 1993.

(Gaizauskas, Humphreys, Wakao, Cunningham 1995), R. Gaizauskas, \linebreak
K.  Humphreys, T. Wakao, H. Cunningham, {\em LaSIE -- a Large-Scale Information
Extraction System}, technical report CS - 95 - 27, Department of Computer
Science, University of Sheffield, 1995.

(Gaizauskas, Humphreys, Wakao, Cunningham 1996), R. Gaizauskas, \linebreak
K.  Humphreys, T. Wakao, H. Cunningham, {\em LaSIE -- Description of the
Sheffield System Used for MUC-6},
in (ARPA 1996).

(Galliers, Sparck Jones 1993) J. Galliers, K. Sparck Jones, {\em Evaluating 
Natural Language Processing Systems}, Technical Report 291, Computer
Laboratory, University of Cambridge.

(Goldfarb 1990) C.F. Goldfarb, {\em The SGML Handbook}, Clarendon Press
1990.

(Grishman 1995) R. Grishman, {\em TIPSTER architecture 1.50, The Tinman
Architecture}, presentation 18/5/95, slides available at \linebreak
http://www.cs.nyu.edu/tipster.

(Grishman, Dunning, Callan 1994)  R. Grishman, T. Dunning, J, Callan,
{\em TIPSTER II Architecture
Design Document Version 1.52 (Tinman Architecture)},
TIPSTER working paper 1995, \linebreak available at
http://www.cs.nyu.edu/tipster.

(Hudson 1995) R. Hudson, GLOSSASOFT: Methods and Guidelines for Software
Internationalisation and Localisation,
{\em Proceedings of the Second Language Engineering
Convention}, London 1995.

(Jacobs ed. 1992) P.S. Jacobs, ed., {\em Intelligent Text-Based Systems,
Current Researh and Practice in Information Extraction and Retrieval},
Lawrence Erlbaum, Hillsdale NJ, 1992.

(Keffer 1995) T. Keffer, {\em Tools.h++ Introduction and Reference Manual},
Rogue Wave 1995.

(Kroll 1994) E. Kroll, {\em The Whole Internet Guide and Catalog, 2nd Edtn.}, 
O'Reilly, Sebastopol CA, 1994.

(Kuno, Oettinger 1962) S. Kuno, A. Oettinger, {\em Multiple Path
Syntaxctic Analyzer}, {\em Information
Processing} 1962 pp 306-312 Proc. IFIP 1962, Popplewell (ed.)
Amsterdam: North Holland.

(Leech, Garside, Atwell 1983) G. Leech, R. Garside, E. Atwell, {\em The
Automatic Grammatical Tagging of the LOB Corpus}, ICAME News 7 1983.

(Lehrberger, Bourbeau 1988) J. Lehrberger, L. Bourbeau, Machine Translation:
linguistic characteristics of MT systems
and general methodology of evaluation. J Benjamins: Amsterdam, 1988.

(Magerman 1994) D. Magerman, {\em Natural Language Parsing as Statistical
Pattern Recognition}, PhD thesis, Department of Computer Science, Stanford,
1994.

(Magerman 1995) D. Magerman, {\em Statistical Decision-Tree Models for
Parsing}, proceedings of ACL 1995.

(Miller, Beckwith, Fellbaum, Gross, Miller 1993) G.A. Miller, R. Beckwith,
C. Fellbaum, D. Gross, K. Miller, Introduction to WordNet: an On-line
Lexical Database, distributed with the WordNet software, 1993.

(Nirenburg, Frederking, Fawell, Wilks 1994)
S. Nirenburg, R. Frederking, D. Fawell, Y. Wilks,
{\em Two Types of Adaptive MT Environments}, proceedings of COLING 94.

(Nyberg, Mitamura, Carbonell 1994)
G.H. Nyberg, T. Mitamura, J.G.Carbonell, {\em
Evaluation Metrics for KBMT}, proceedings of COLING 1994.

(Onyshkevych 1995a) B. Onyshkevych, {\em Information Extraction}
\linebreak {\em 
Task Definition}, MUC-6 documentation, \linebreak
http://cs.nyu.edu/cs/faculty/grishman/muc6.html, 1995.

(Onyshkevych 1995b) B. Onyshkevych, {\em Template Sketches for Scenarios},
MUC-6 documentation,
http://cs.nyu.edu/cs/faculty/grishman/muc6.html,
1995.

(Ousterhout 1994) J.K. Ousterhout, {\em Tcl and the Tk Toolkit},
Addison-Wesley 1994.

(Palmer, Finin 1990) M. Palmer, T. Finin, {\em Workshop on the Evaluation of
Natural Language Processing Systems}, {\em Computational Linguistics}, Vol.
16, Number 3.

(Robert, Armstrong 1995) G. Robert, S. Armstrong, {\em Tagging Tool},
LRE 62-050 Deliverable 2.4.1, 1995.

(Rocca, Black, Cunningham, Zarri, Celnik 1993)
G. Rocca, W.J. Black, H. Cunningham, P. Zarri, P. Celnik, {\em COBALT
-- COnstruction, augmentation and use of knowledge BAses
from natural Language documenTs}, LRE Project 61-011 Deliverable
2, 1993.

(Salton 1989) G. Salton, Automatic Text Processing: 
The Transformation, Analysis, and Retrieval of Information by Computer,
Addison-Wesley, Reading MA, 1989.

(Sch\"{u}tz 1994) J. Sch\"{u}tz, {\em Developing Lingware in ALEP}, in the
{ALEP User Group News}, Issue 1, CEC Luxemburg, October 1994.

(Simpkins 1992) ALEP User Guide, CEC Luxemburg, 1992.

(Sparck Jones 1994) K. Sparck Jones, {\em Towards better NLP System
Evaluation}, Proceedings 2nd ARPA Workshop on Human Language Technology.

(Strzalkowski, Scheyen 1993) T. Strzalkowski, P.G.N. Scheyen, {\em
Evaluation of TTP Parser: A Preliminary Report}, PROTEUS Project Memorandum
\#58, Dept. Computer Science, New York University 1993.

(Sundheim 1995a) B. Sundheim, {\em Tokenization Rules},
MUC-6 documentation,
http://cs.nyu.edu/cs/faculty/grishman/muc6.html,
1995.

(Sundheim 1995b) B. Sundheim, {\em Named Entity Task Definition},
MUC-6 documentation,
http://cs.nyu.edu/cs/faculty/grishman/muc6.html,
1995.

(Thompson 1995a) H. Thompson, {\em Priority On-Ramp or Squashed Hedgehog?},
invited talk, {\em Second Language Engineering Conference}, London 1995.

(Thompson 1995b) H. Thompson, {\em MULTEXT Workpackage 2 Milestone B
Deliverable Overview}, LRE 62-050 Deliverable 2, 1995.

(TIPSTER 1994) TIPSTER Architecture Committee, {\em TIPSTER Text Phase II
Architecture Concept}, 
TIPSTER working paper 1994, available at
\linebreak
http://www.cs.nyu.edu/tipster.

(Wakao, Gaizauskas, 1995), T. Wakao, R. Gaizauskas,
{\em Evaluation of Algorithms for the Recoginition and Classification
of Proper Names}, {\em submitted}.

(Wilks 1994) Y. Wilks, {\em Developments in MT Research in the US}, ASLIB
Proceedings, Vol. 46 \#4, April 1994.

(Wilks, Gaizauskas 1994) Y. Wilks, R.G. Gaizauskas, {\em A Research Proposal
on Large-Scale Information Extraction}, EPSRC proposal October 1994.

(Wilks, Guthrie, Guthrie, Cowie 1992)
Y. Wilks, L. Guthrie, J. Guthrie, J. Cowie, {\em
Combining Weak Methods in Large-Scale Text Processing}, in (Jacobs ed.
1992).

(Wilks, Guthrie, Slator 1996) Y. Wilks, L. Guthrie, B. Slator, {\em Electric
Words}, MIT Press, Cambridge, MA 1996.

(Winograd 1972) T. Winograd, {\em Understanding Natural Language}, Academic
Press, New York 1972.

\section*{APPENDICES}

\appendix

\section{\label{related_work}Related work}

We discuss three systems here, ALEP, MULTEXT and the TIPSTER Architecture.

\subsubsection*{ALEP}

ALEP (Simpkins 1992) is an EC project which aims to provide an
Advanced Language Engineering Platform -- superficially a similar goal to
ours. The approaches are quite different, however. ALEP is an advanced
system for developing and manipulating feature structure knowledge-bases
under unification. Also provided are several parsing algorithms, algorithms
for transfer, synthesis and generation (Sch\"{u}tz 1994).
As such it is a system for developing particular types of data resource and
for {\em doing} a particular set of tasks
in LE in a particular way.
ALEP does not aim for complete genericity (or it would be in the business of
providing algorithms for Baum-Welch estimation, or fast regular expression
matching, or \ldots).
Supplying a generic system to do every LE task is clearly
impossible, and prone to instant obsolescence in a rapidly changing field.
GATE, in contrast, is a shell, a backplane into which the whole spectrum of LE
modules and databases can be plugged.
Components used within GATE will typically exist already -- our emphasis is
reuse, not reimplementation.
Our project is to provide a flexible and efficient
way to combine LE components to make LE systems (whether experimental or for
delivered applications) -- not to provide `the one true system', or even
`the one true development environment':
ALEP-based systems might well provide components operating within GATE.

The ALEP enterprise, then, is orthogonal to ours --  there is no significant
overlap or conflict.

\subsubsection*{MULTEXT}

MULTEXT (Ballim 1995; Thompson 1995b; Finch, Thompson, McKelvie 1995) 
is another EC project, whose
objective is to produce tools for multilingual corpus
annotation and sample corpora marked-up according to the same standards used
to drive the tool development. Annotation tools under development perform
text segmentation, 
POS tagging, morphological analysis and parallel text alignment.
The project has defined an architecture centred on a
model of the data passed between the various phases of processing implemented
by the tools.

MULTEXT is based on SGML, the Standard Generalised Markup Language (Goldfarb
1990). SGML works by adding extra information to texts in a standard format.
For example, the (rather short) news article

Reuter \\
Dog bites man. \\
Newshound implicated.

might appear in SGML as

$<$DOC$>$ \\
$<$HEADERS$>$Reuter$<$/HEADERS$>$ \\
$<$SENT$>$Dog bites man.$<$/SENT$>$ \\
$<$SENT$>$Newshound implicated.$<$/SENT$>$ \\
$<$/DOC$>$

Markup is between chevrons, `$<$' and `$>$'; slashes signify the end of a
marked-up entity. The language is information-neutral (the tags `DOC',
`SENT' etc. are not part of the language definition) and is encoded in
whatever character set the source text originates in (e.g. ASCII).

The MULTEXT architecture is based on a commitment to TEI-style
(the Text Encoding Initiative (Sperberg-McQueen, Burnard 1994))
SGML encoding of information about text. The TEI defines standard tag sets
for a range of purposes including many relevant to LE systems.
Tools in a MULTEXT system communicate
via interfaces specified as SGML document type definitions (DTDs --
essentially tag set descriptions),
using character streams on pipes -- an arrangement modelled
after UNIX-style shell programming. This UNIX flavour was also apparent in
provision for record/field encoding (records $=$ lines of text; fields $=$
whitespace-separated character groups) of markup as an interchangeable
alternative to SGML (though this was dropped from later versions of the
tools), and in provision of an SGML-aware version of {\tt sed}, the UNIX
pattern search and replace tool. Organisational problems have,
unfortunately, led to an early termination of the project, but the tools and
the architecture they run in will still be completed and distributed for
research purposes.

MULTEXT endorses the view that SGML is an appropriate
and flexible language for the 
splitting and recombination of text analysis elements.
A tool selects what information it requires from its input SGML
stream and adds information as new SGML markup. An advantage here is a
degree of data-structure independence: so long as the necessary
information is present in its input, a tool can ignore changes to other
markup that inhabits the same stream -- unknown SGML is simply
passed through unchanged. A disadvantage is that although graph-structured
data may be expressed in SGML, doing so is complex (either via concurrent
markup, the specification of multiple legal markup trees in the DTD, or by
rather ugly nesting tricks to cope with overlapping, aka ``milestone tags'').
Graph-structured information might be present in the output of a parser, for
example, representing competing analyses of areas of text.

Another feature of MULTEXT is a set of abstract data types (ADTs)
for all tool I/O (Ballim 1995) supported by a single shared API
(Application Program(mers') Interface)
for access to the types. An executive (the {\em tool shell}) glues tools
together in particular configurations according to user specifiactions. The
shell may extract sub-trees from SGML documents to reduce the I/O load where
tools only require a subset of a marked-up document.

The ADT set forms an object-oriented model%
\footnote{OO in the sense of using inheritance and data encapsulation.}
of the data present in a marked-up document. Example classes include {\bf
Sentence}, {\bf SentenceBlock} (sequence of sentences), {\bf LexicalWord}
(word plus definition from a lexicon). The ADT model reflects the type of
processing available in the tool set --
there is a type {\bf TaggedSentence}, for
example, but not a {\bf ParsedSentence}.

Finally, MULTEXT has developed some general support infrastructure for
handling SGML and for parallelising tool pipelines.
A query language for accessing components of SGML
documents is defined 
and API in support of this language provided. For example
a program might specify parts of a document by the pattern
{\tt DOC/*/s}
which refers to all \verb|<s>| objects under \verb|<DOC>| tags -- all
SGML-marked sentences in the document.
Additionally SGML-aware versions of various UNIX utilities are in
development. Parallel execution may be supported at the level of single tools
via a program that distributes pipelined operations over a set of networked
machines.

MULTEXT is implemented for the UNIX platform. Access to tools is as unitary
programs and via the tool shell; the SGML query language is supported by a C
API. The consortium has declared an
intention to make implementations generally available, and although the
project is finishing early due to logistic difficulties, most tools and the
support shell will continue development at ISSCO for release early in 1996.

Summary:
\bit
\item
MULTEXT tools operate on SGML streams. 
\item
An object model of the data
in those streams is defined, along with
\item
an API to access the data.
\item
An API
and query language for accessing components of SGML documents is provided
along with
\item
various useful SGML-aware tools.
\eit
\resetparskip

\subsubsection*{\label{tipster}TIPSTER II}

The TIPSTER programme in the US, currently in its second phase, has also
produced a data-driven architecture for NLP systems (Grishman, Dunning,
Callan 1995; TIPSTER Architecture Committee 1994). Like MULTEXT, TIPSTER
addresses specific forms of language processing, in this case information
extraction and document detection (or information retrieval -- IR). As will
become clear below, however, TIPSTER's approach is not restricted to
particular NL tasks.

Whereas in MULTEXT all information about a text is encoded in SGML, 
which is added by the tools, in
TIPSTER a text remains unchanged while information is stored in a separate
database in the form of {\em annotations}. Annotations associate portions of
documents (identified by sets of start/end byte offsets or {\em spans})

with
analysis information ({\em attributes}), e.g.: POS tags;  textual unit type;
template element. In this way the information built up about a text by NLP
(or IR) modules is kept separate from the texts themselves. In place of an
SGML DTD an {\em annotation type declaration} defines the information
present in annotation sets, for example a set of values for MUC-style
organisation template elements. Figure \ref{annotations_eg} shows an example
from (Grishman, Dunning, Callan 1995). SGML I/O is catered for by API calls
to import and export SGML-encoded text.

\small
\begin{figure}[!htbp]

\small

\begin{center}   
\begin{tabular}{|l|l|r|r|l|} \hline
\multicolumn{5}{|c|}{\em Text} \\
\multicolumn{5}{|c|}{{\tt Sarah savored the soup.}} \\
\multicolumn{5}{|c|}{{\tt 0...|5...|10..|15..|20}} \\ \hline \hline
\multicolumn{5}{|c|}{\em Annotations} \\
Id & Type  & \multicolumn{2}{c|}{Span} & Attributes \\
   &       & Start & End &                          \\ \hline
 1 & token & 0  & 5  & pos=NP \\
 2 & token & 6  & 13 & pos=VBD \\
 3 & token & 14 & 17 & pos=DT \\
 4 & token & 18 & 22 & pos=NN \\
 5 & token & 22 & 23 &  \\ \hline
 6 & name  & 0  & 5  & name\_type=person \\ \hline
 7 & sentence & 0 & 23 & \\ \hline
\end{tabular}
\end{center}

\caption{TIPSTER annotations example \label{annotations_eg}}
\end{figure}
\normalsize

The definition of annotations in TIPSTER forms part of an object-oriented
model that deals with inter-textual information as well as single texts.
Documents are grouped into {\em collections}, each with a database storing
annotations and document attributes such as identifiers, headlines etc.
Collections are the first-class entities in the architecture. The
model also describes elements of IE and IR systems relating to their use,
with classes representing queries and information needs.

The TIPSTER architecture is designed to be portable to a range of operating
environments, so it does not define implementation technologies. Particular
implementations make their own decisions regarding issues such as
parallelism, user interface, or delivery platform.
An implementation in C and Tcl (Ousterhoot 1994) from CRL
(the Computing Research Lab, New Mexico State University)
implements client-server operation (using Tcl-dp), a server database manager
fielding requests from
client modules.

This implementation is available now and includes both C and
Tcl APIs. It is not currently portable beyond UNIX, though Tcl/Tk 
is becoming available on Windows and Macintosh.

The architecture was the result of unpaid collaboration between a large
number of ARPA-supported sites in the US.

\subsubsection*{Comparison of MULTEXT and TIPSTER}

Both projects propose architectures appropriate for LE, but there are a
number of significant differences. We discuss seven here, then note the
possibility of complimentary interoperation of the two.

\ben
\item
MULTEXT adds new information to documents
by augmenting an SGML stream; TIPSTER stores information remotely in a
dedicated database. This has several implications. Firstly, TIPSTER can
support documents on read-only media (e.g. CD-ROMs, which may be used for
bulk storage by organisations with large archiving needs, even though access
will then be slower than from hard disk). Secondly, TIPSTER
avoids the difficulties referred to earlier of representing graph-structured
information in SGML. From the point of view of efficiency, the original
MULTEXT model of interposing SGML between all modules implies a generation
and parsing overhead in each module. Later versions have replaced this model
with a pre-parsed representation of SGML to reduce this overhead. This 
representation will presumably be stored in intermediate files, which
implies an overhead from the I/O involved in continually reading and writing
all the data associated with a document to file. There would seem no reason
why these files should not be replaced by a database implementation,
however, with potential performance benefits from the ability to do I/O on
subsets of information about documents (and from the high level of
optimisation present in modern database technology).

\item
A related issue is storage overhead. TIPSTER is minimal in this respect, as
there is no inherent need to duplicate the source text (which also means
that it works naturally with read-only media like CD-ROMs). MULTEXT
potentially has to duplicate the source text at each intermediary stage,
although this might be ameliorated by shifting to a database implementation.

\item
TIPSTER's data architecture is process-neutral -- the objects in the model
are generic to all information that is associated with definite ranges
of text. (The more concrete aspects of the architecture to do with IE and
IR model the objects involved in user interaction with such systems.)
MULTEXT's model is tool-specific, as noted above (although the underlying
representation language, SGML, is information-neutral).

\item
There is no easy way in an SGML-based system to differentiate sets of
results (i.e. sets of markup) by e.g. the program or user that originated
them. In general, storing information about the information present in an
SGML system (or {\em meta-information}) is messy. This is a problem for
MULTEXT but not for TIPSTER. A related point is that TIPSTER can easily
support multi-level access control via a database's protection mechanisms --
this is again not straightforward in SGML.

\item
Distributed control is easy to implement in a database-centred system like
TIPSTER -- the DB can act as a blackboard, and implementations can take
advantage of well-understood access control (locking) technology. How to do
distributed control in MULTEXT is not obvious.

\item
TIPSTER provides no tools or databases, but many sites are already committed
to TIPSTER-compatibility, so the set of modules available in the framework
will grow over time.
MULTEXT is based
around a set of tools and reference corpora annotated accordingly.

\item
Working implementations of TIPSTER have been available for some months now;
MULTEXT will be distributed in 1996.
\een
Interestingly, a TIPSTER system could function as a module in a MULTEXT
system, or vice-versa. A TIPSTER storage system could write data in SGML for
processing by MULTEXT tools, and convert the SGML results back into native
format. Also, the extensive work done on SGML processing in MULTEXT could
usefully fill a gap in the current TIPSTER model, in which SGML capability
is not fully specified (plans are currently being formed in the US
to address this
problem -- input from European experience would seem advisable).
Integration of the results of both projects would
seem to be the best of both worlds, and we hope to achieve this in GATE.
\resetparskip

Note that we believe that 
SGML and the TEI must remain central to any serious text
processing strategy. The points above do not contradict this view, but
indicate that SGML should not form the central representation format of every
text processing system. Input from SGML text and TEI conformant output are
becoming increasingly necessary for LE applications as more and more
publishing adopts these standards. This does not mean, however, that
flat-file SGML is an appropriate format for an architecture for LE systems.
This observation is born out by the fact that
TIPSTER started with an SGML/TEI architecture but rejected it in favour of
the current database model, and that MULTEXT has gone halfway to this style by
passing pre-parsed SGML between components.

\section{\label{gate_design}GATE -- design and implementation}

{\bf Note:} this appendix is a preliminary version of design and 
implementation
documentation for GATE and VIE. It is a) incomplete and speculative,
and b) repeats some material from earlier sections of the report.

\subsection*{Architecture overview}

GATE is based on a combination of the TIPSTER and MULTEXT models. The
centrepiece of the architecture is a TIPSTER-style document management
database, chosen for reasons of efficiency, maturity of implementation and
ease of extensibility. The current TIPSTER model provides hooks for the
incorporation of SGML support but has not fully developed this aspect of
the architecture (see above).
We plan to capitalise on the work done in MULTEXT to
augment the SGML capabilities of the TIPSTER architecture, probably via the
development of a unified API. This unification will not be available in the
initial release of GATE (early 1996), however.

GATE will form a bridge between the American work and the European work on
SGML and conformance to the TEI guidelines and DTDs.

\subsubsection*{Components}

\begin{figure}[!htbp]
\centerline{\psfig{figure=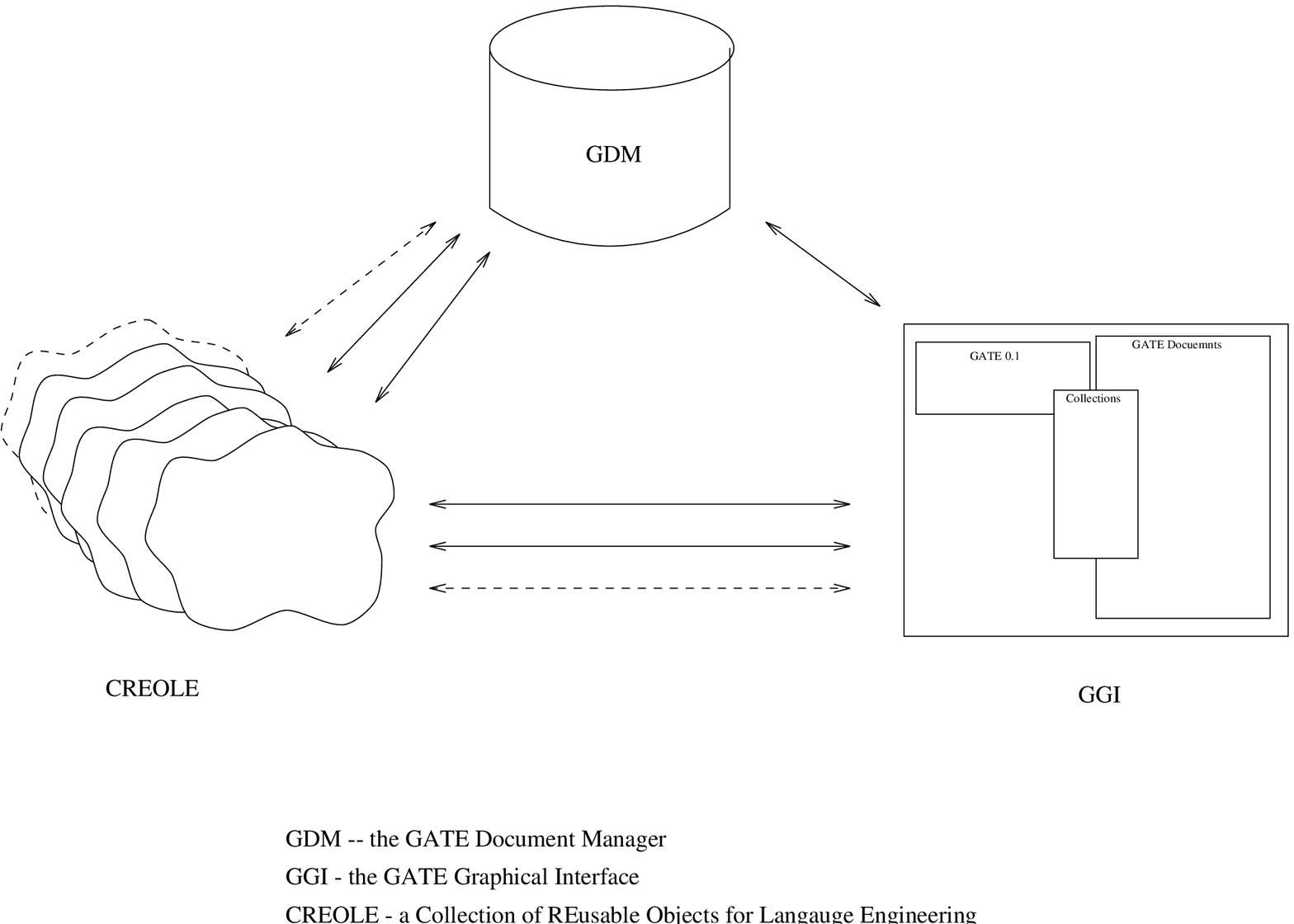,height=8cm}}
\caption{\label{arch}The three elements of GATE}
\end{figure}

GATE comprises three principal components (see figure \ref{arch}):
\bde
\item[GDM --]
the GATE document manager, based on the
TIPSTER document manager with added SGML
capabilities (and using an implementation from CRL at NMSU, whose assistance
we gratefully acknowledge);
\item[GGI --]
the GATE graphical interface, a development tool for LE R\&D, providing
integrated access to the services of the other components and adding
visualisation and debugging tools;
\item[CREOLE --]
a Collection of REusable Objects for Language Engineering: the set of
modules integrated with the system. CREOLE comprises wrappers for existing
modules, which may or may not require changing, plus modules developed
explicitly for GATE compliance. Some objects are process-orientated, some
data-oriented.
\ede
The first distribution of GATE will be configured as a support tool for
collaborative R\&D in Information Extraction by the inclusion of a
CREOLE set that implements a full-scale
MUC-compatible IE system (called VIE, the Vanilla IE
system -- see section \ref{vie}).
\resetparskip

MULTEXT integration will involve:
\bit
\item
creating CREOLE object wrappers for the tool set;
\item
providing SGML I/O and SGML manipulation via and API based on the MULTEXT
work.
\eit
\resetparskip

\subsection*{\label{integration}Integrating CREOLE objects}

As noted above, GATE is {\em not} a system for doing LE, but a backplane
to assemble processing modules and databases to form LE systems
(whether
experimental or for end-user delivery). The analogy here is with extensible
computer hardware architectures -- expansion cards in a PC, for example.
Just as producing a card to do fast video off a VESA bus or to drive a serial
line from an ISA slot means conforming to the protocols defined by those
architectures, so integrating LE objects in GATE 
(i.e. producing a new member of CREOLE) imposes some
interface constraints. These constraints are in the form of
functions that must be available for GDM and GGI, and are described here.

When the user initiates a particular CREOLE object via GGI (or when a
programmer does the same via the GATE API when building an LE application)
the object is initialised using the standard calls provided in the CREOLE
wrapper. The object then runs, obtaining the information it needs (document
source, annotations from other objects) via calls to the GDM API. Its
results are then stored in the GDM database and become available for
examination via GGI or to be the input to other CREOLE objects.

\begin{figure}[!htbp]
  \centerline{\psfig{figure=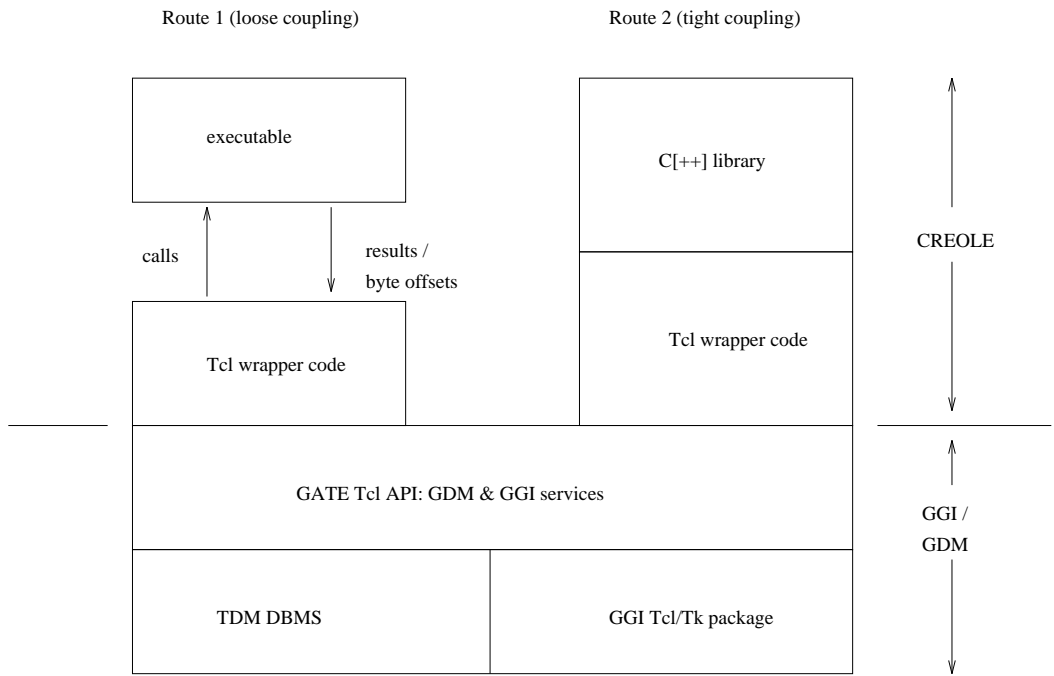,height=9cm}}
  \caption{\label{creole_eg}CREOLE integration routes}
\end{figure}

Figure \ref{creole_eg} shows the two ways to provide the CREOLE wrapper
functions. Packages written in C or in languages which can be used as
libraries with C linkage conventions can be compiled into GATE directly as a
Tcl package (see Ousterhout 1994 chapter 31). This is {\em tight
coupling} (route 2 in the diagram).
Alternatively the
underlying implementation of services can be via an executable ({\em loose
coupling}, route 1). This
executable is then called by the CREOLE wrapper code. In either case the
implementation of CREOLE services is completely transparent to GATE.

CREOLE wrappers encapsulate information about the preconditions for a
module to run (data that must be present in the GDM database) and
post-conditions (data that will result). This information is needed by GGI
-- see below.
Note that aside from the information needed for GGI to provide access to a
module, GATE compatability equals TIPSTER compatability -- i.e. there will
be very little overhead in making any TIPSTER module run in GATE.

In addition to the macro requirements on CREOLE integration described above,
GDM imposes constraints on the I/O format of CREOLE objects, namely that all
information must be associated with byte offsets and conform to the
annotations model of the TIPSTER architecture (see appendix 
\ref{related_work}).
The principal overhead in this process is making the components being
integrated use byte offsets, if they don't already do so. Where components
use SGML, I/O filters will convert markup to the TIPSTER style.

As we noted above CREOLE objects may be data-orientated. It is our intention
to integrate as large a set of LE data resources as possible within GATE in
order to reduce the overhead of installing and understanding the software
interfaces of these resources. For example, the Wordnet
thesaurus (Miller, Beckwith, Fellbaum, Gross, Miller 1993) will be
given a CREOLE wrapper encapsulating the C API as a GATE service. Grammars,
lexica, gazetteers -- all are candidates for CREOLE integration, and as the
set expands GATE can become a standard resource repository for LE data as
well as LE processing modules.

\subsection*{GGI}

GGI is a graphical tool that encapsulates the GDM and CREOLE resources in a
fashion suitable for interactive building and testing of LE components and
systems. The philosophy is to provide a rich set of tools including but not
limited to the CREOLE modules. So, for example, access to a KWIC tool
or the WordNet interface is included, as well as taggers, parsers, etc.
from CREOLE.

GGI is intended for developers. Delivered systems built on GATE will not 
generally use GGI (though they may be able to reuse parts of the interface
for their own front-ends).

GGI adopts the OSF Motif look and feel, provided via the Tcl/Tk toolkit (as
used, for example, in Netscape).%
\footnote{Tcl/Tk will be available in native look and feel for PC/Windows
and Macintosh some time in 1996, so GATE may at that point be able to
migrate to these platforms.}

\subsubsection*{GGI}

The current version of the interface is has gone through several redesign
iterations based on feedback on initial prototypes.

\begin{figure}[!htbp]
  \centerline{\psfig{figure=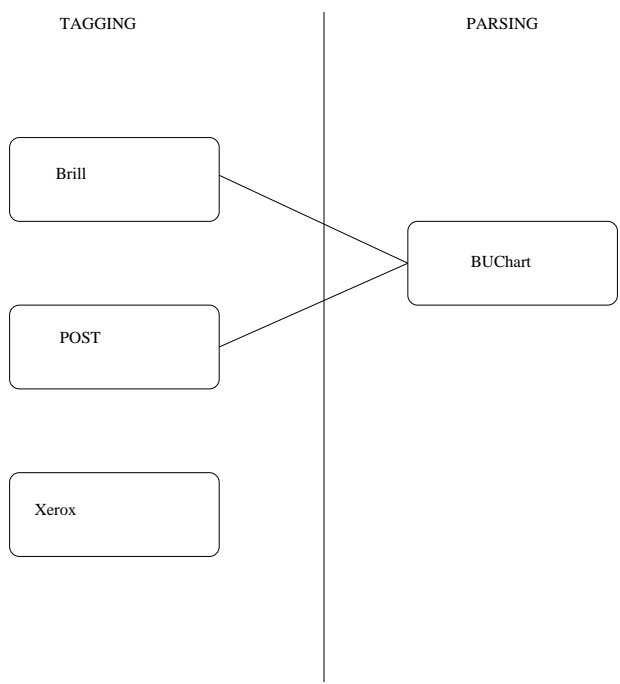,height=9cm}}
  \caption{\label{ggi-0.3}GGI objects graph example}
\end{figure}

Launching CREOLE processes is done via
a partially connected graph of possible paths through the
processes embodied by the systems and modules menus of 0.1.%
\footnote{Thanks particularly to Kevin Humphreys for this idea.}
The idea is that each module that is applicable to the LE task under
development (IE, MT, \ldots) is given a button in a large canvas window.
Clicking the button will run the process associated with the button. Figure
\ref{ggi-0.3} shows a small example. Here we have a choice of whether to run
the Brill or POST taggers, both of which may produce results required by the
BUChart parser, or the Xerox tagger which will not produce results
appropriate to the parser.

Interestingly, this arrangement for viewing and launching chains of LE
modules quite closely parallels the {\em braided evaluation} model proposed
in (Crouch, Gaizauskas, Netter 1995). This suggests that
implementation of the
evaluation schemes discussed there and in (Galliers, Sparck Jones 1993;
Sparck Jones 1994) might be facilitated by GATE.

The first task facing a user is to open a collection.
It is not mandatory to then open a specific document -- functions may be
run on whole collections as well as single documents (though it may be
appropriate to have a warning dialog box as large batched runs ``may take a
little while''). 
There is a file menu in the top left corner, with {\tt open
collection} leading to a list of collections, followed by a list of
documents.

This is a view of the system as a set of processes that may be linked in
pipelines in various ways. The permissible paths through the graph depend on
the data that a module requires as its input. GATE systems are built from
combinations of modules chained together. These chains may be represented by
highlighted paths through the graph (e.g. selecting LaSIE from a systems
menu would highlight the arcs connecting the various LaSIE object nodes).
Clicking a module within a chain will
then run all those in the remaining portion of the chain.

Visual information regarding the data present in the system is represented
in two ways: by colour-coding of the module buttons and by colour-coding of
the result (with an associated colour key).

The module buttons change colour depending on whether the data that the
process produces is present or not:
\bde
\item[green] ready to run, data not present;
\item[amber] requires data from a previous stage to run;
\item[red] data available (process has already been run successfully on
current document / collection.
\ede
Clicking on a green button runs the relevant CREOLE object (via a pop-up
dialog if options need setting);
clicking on amber generates a menu of possible preceding modules to run;
click red and a menu of results to view is displayed.
\resetparskip

An optional pop-up launched from the file menu displays the
output of modules as they run.

Given that the set of CREOLE modules will be large it will be necessary to 
allow a large screen space for the graph, and for it to be X and Y scrollable.
Perhaps processing stages should be collapsible. Note that the
implementation of the display will be non-trivial, and will require the use
of some drawing algorithm like that used in daVinci (Fr\"{o}lich, Werner
1994).

Further information regarding the data produced by CREOLE objects is
delivered by colour-coded displays of documents, e.g. a text might be
displayed with coreference chains displayed in green. Each type of result
also specifies a colour key, to be displayed on a bar with the result
viewer.

The implementation of the processes graph should be via configuration
information supplied with each CREOLE wrapper -- i.e. there should be no
information hard-coded into GATE regarding different modules. This might be
achieved for example by each object registering
its name, version and result type.%
\footnote{Each object then also specifies a set of
preconditions in the form of regular expressions matching these annotations,
e.g. the Brill tagger might store
\bit
\item brill-0.1 pos\_tags
\eit
and a parser that required the tags to run might then specify
\bit
\item brill-* pos\_tags, or
\item * pos\_tags, or
\item (brill)|(post)-* pos\_tags.
\eit}\
It should also be possible to specify standard ways for results to be
displayed (via an annotation type/colour key table, for example). These
details need more work.
\resetparskip

\subsection*{Implementation technology}

GATE is implemented in a mixture of Tcl/Tk and C[++]. The glue between the
various components is Tcl, a script language developed specifically for
systems integration (Ousterhout 1994). In common with other script languages,
like Perl or the Bourne shell plus UNIX utilities, Tcl provides high-level
constructs and facilities that greatly simplify the implementation of simple
systems.
Unlike other script languages, Tcl also has an extremely clean C interface,
allowing seamless integration of C[++] libraries with Tcl scripts.
GATE's own code, then, is Tcl or C[++] (though this in no way restricts the
implementation technology used in CREOLE modules -- see above).

Another reason for choosing Tcl is the Tk package that comes with it. Tk is
a Tcl library that encapsulates the MOTIF X-Windows toolkit. Whereas
programming X via C is a black art that has spawned legions of expensive and
complex screen-painting utilities, scripting Tk is a simple interactive
process. The initial GGI prototype was coded in less than a week by a novice
Tk programmer.

\tickle are public domain and are under active development by Sun
Microsystems. Forthcoming changes include cross-platform portability across
UNIX/X, MS-Windows and Macintosh.

The GDM API, then, is a set of Tcl calls. These calls are generally also
available in \cpp.

All GATE systems are 8-bit clean, and may therefore be used with languages
that can be represented by 1 byte character sets. Multi-byte character
support is highly desirable (probably via the UNICODE standard). A route
to 16 bit capability might be via a replacement for the Tcl string functions
(maybe using the Tools.h++ library (Keffer 1995))  and via
reimplementation of the Tk text widget (or integration of CRL's Motif
widget).

\end{document}